\documentclass[letterpaper, 10 pt, conference]{ieeeconf}  

\IEEEoverridecommandlockouts                              

\usepackage{balance}
\usepackage{graphicx} 
\usepackage{amsmath} 
\usepackage{amssymb}  
\usepackage{algorithm}
\usepackage{algorithmic}
\usepackage{booktabs}
\title{\LARGE \bf
TacMamba: A Tactile History Compression Adapter Bridging Fast Reflexes and Slow VLA Reasoning
}

\author{Zhenan Wang$^{1,2,\dagger}$, Yanzhe Wang$^{1,\dagger}$, Meixuan Ren$^{1,3}$, Peng Li$^{2}$, Yang Liu$^{1}$, \\
Yifei Nie$^{1,2}$, Limin Long$^{2}$, Yun Ye$^{2}$, Xiaofeng Wang$^{2}$, Zhen Zhu$^{2}$ and Huixu Dong$^{1,*}$
\thanks{$^{\dagger}$These authors contributed equally to this work.}%
\thanks{$^{*}$Corresponding author.}%
\thanks{$^{1}$ Zhejiang University, Hangzhou, China}%
\thanks{$^{2}$ GigaAI, Beijing, China }%
\thanks{$^{3}$ Harbin Institute of Technology, Harbin, China }%
}

\begin{document}

\maketitle
\thispagestyle{empty}
\pagestyle{empty}

\begin{figure*}
    \includegraphics[width=1.0\linewidth]{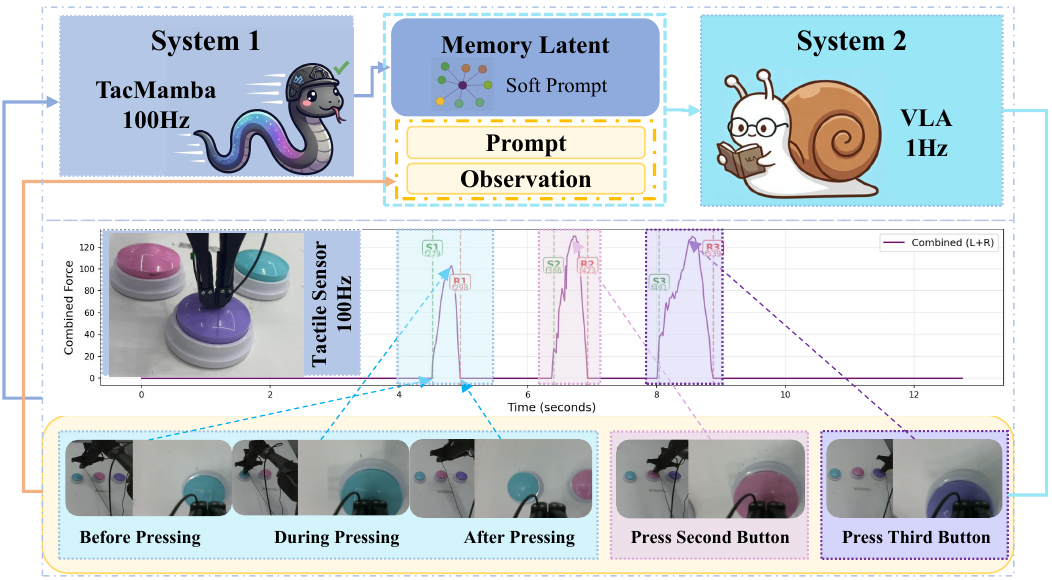} 
    \caption{\textbf{The TacMamba System Architecture.} The framework bridges the spatiotemporal discrepancy between high-speed reflexes and low-frequency reasoning. \textbf{Left (System 1):} The tactile encoder processes 1D force streams at 100Hz using Mamba Models, recursively updating the hidden state $h_t$ in real-time. \textbf{Right (System 2):} This compressed hidden state $h_t$ is projected and asynchronously injected as a soft prompt into the low-frequency ($\sim$1Hz) Vision-Language-Action (VLA) planner.}
    \label{fig:system_arch}
\end{figure*}

\begin{abstract}
In visually ambiguous manipulation such as detecting button's ``click'' tactile feedback is often the sole source of ground truth. However, fusing tactile data poses a significant challenge due to a \textbf{spatiotemporal mismatch}: tactile perception requires high-frequency processing with long-horizon memory (System 1), whereas visual policies operate at low control frequencies (System 2). Existing architectures struggle to bridge this gap: Transformers are computationally prohibitive for high-frequency loops ($>100$Hz), while LSTMs suffer from forgetting over extended interaction histories. In this paper, we introduce \textbf{TacMamba}, a hierarchical architecture that aligns high-bandwidth tactile reflexes with low-frequency visual planning. Our approach comprises three core contributions: (1) a custom high-frequency tactile interface designed for flexible integration; (2) a Mamba-based \textbf{Tactile History Compressor} that encodes continuous force history into a compact state with \textbf{$\mathcal{O}(1)$ inference latency (0.45 ms)}, enabling plug-and-play fusion with VLA models without joint pre-training and (3) a \textbf{Tactile-Guided Dual-Stage Training} strategy that leverages temporal discrimination for self-supervised representation learning and phase-uniform sampling to mitigate data sparsity. Experiments on discrete counting and implicit state switching demonstrate that TacMamba achieves \textbf{100\% success rates}, significantly outperforming the visual-only $\pi_{0.5}$ baseline, while strictly satisfying hard real-time constraints.
\end{abstract}

\section{INTRODUCTION}

Human dexterity synergizes slow, semantic planning (System 2) with fast, reflexive reactions (System 1)~\cite{kahneman2011thinking}. While recent Vision-Language-Action (VLA) models have demonstrated proficiency in System 2 capabilities~\cite{black2025pi,rt2_2023,octo_2023,team2025gigabrain,zhai2025igniting,kim2025fine}, they struggle to emulate the reflexive nature of System 1. The fundamental bottleneck lies in a frequency mismatch: existing architectures cannot efficiently process high-frequency tactile history, creating a disconnect between high-level visual reasoning and low-level physical interaction.

This discrepancy is critical in scenarios characterized by visual ambiguity, such as detecting a subtle button ``click'' or identifying stripped threads~\cite{cheng2025omnivtla,yu2025forcevla,yang2024binding}. In such cases, tactile feedback serves as the indispensable ground truth. Unlike vision, which captures discrete snapshots, tactile perception relies on the temporal profile of contact---continuous force curves and transient vibrations. Consequently, robust manipulation necessitates a model with long-term temporal memory to distinguish successful actuation from failure based on interaction history rather than instantaneous state~\cite{Shi2025MemoryVLAPM,li2025map}.

However, modeling this extended history imposes a prohibitive computational burden. High-resolution sensors stream data at frequencies exceeding 100Hz~\cite{yuan2017gelsight,lambeta2020digit}. Processing merely two seconds of history with standard Transformers generates hundreds of tokens~\cite{vaswani2017attention}, causing computational costs to scale quadratically ($\mathcal{O}(N^2)$). This latency bottlenecks the high-frequency control loop, effectively limiting the ``Fast'' system to the response time of the ``Slow'' visual planner. Conversely, traditional RNNs (e.g., LSTMs) offer fast inference but suffer from catastrophic forgetting, failing to retain fine-grained physical details over thousands of time steps~\cite{pascanu2013difficulty,bai2018empirical,khandelwal18context}.

Beyond architectural limits, multi-modal training poses two severe bottlenecks. First, integrating high-frequency tactile streams into massive VLAs via end-to-end co-training is computationally prohibitive~\cite{rt2_2023, black2025pi} and risks catastrophic forgetting of established semantic priors~\cite{yu2025forcevla, cheng2025omnivtla}. Second, contact-rich tasks suffer from extreme temporal data imbalance~\cite{fang2025kalm}. Critical interaction events, such as button ``snap-throughs'' (Fig.~\ref{fig:system_arch}), are highly transient ($<5\%$ of a trajectory)~\cite{johns2021coarse}. Standard uniform sampling drowns these vital signals in trivial reaching motions, leading to policy failures precisely at the moment of contact.

To resolve these spatiotemporal and data challenges, we introduce \textbf{TacMamba}, a unified hardware-software framework. As illustrated in Fig.~\ref{fig:system_arch}, TacMamba bridges the frequency gap through an asynchronous architecture: the tactile encoder (System 1) performs continuous streaming inference at 100Hz to update a compressed hidden vector of tactile history, which is then queried on-demand by the low-frequency ($\sim$1Hz) VLA planner (System 2). Our contributions are three fold:\\
\noindent (1) We propose TacMamba, the first hierarchical architecture that aligns high-bandwidth tactile reflexes (100Hz) with low-frequency visual reasoning ($\sim$1Hz) without downsampling. Unlike prior methods that discard high-frequency data, our design enables lossless, real-time processing of micro-scale contact dynamics.\\
(2) A novel Mamba-based tactile encoder~\cite{gu2024mamba,gu2022efficiently} that resolves the accuracy-latency trade-off. By achieving constant-time $\mathcal{O}(1)$ inference without sacrificing perception fidelity, TacMamba significantly outperforms the Transformer baselines in speed. This represents the first architecture capable of processing ultra-long tactile histories within the hard real-time constraints of embedded hardware.\\
(3) We propose a robust plug-and-play two-stage learning framework that circumvents the need for expensive VLA joint pre-training. This comprises: (i) self-supervised pre-training via Ternary Temporal Discrimination to independently capture causal contact dynamics, and (ii) efficient VLA fine-tuning augmented with a novel Phase-Uniform Sampling strategy. This approach resolves the sparsity of critical interaction frames and seamlessly injects tactile history into the VLA, improving data efficiency multiple times.
\section{RELATED WORK}
\subsection{Temporal model in VLA}
Recent advances in VLA models have demonstrated remarkable success in equipping robots with high-level semantic reasoning capabilities (System 2)~\cite{rt2_2023,octo_2023,black2025pi}. Building upon foundation models, recent work has further enhanced the generalization of VLAs in open-world scenarios~\cite{team2025gigabrain,zhai2025igniting}. Furthermore, to address the challenge of state aliasing in long-horizon tasks, recent frameworks have integrated memory mechanisms. For instance, MAP-VLA~\cite{li2025map} constructs a memory bank from historical demonstrations using retrieval-augmented prompting, while MemoryVLA~\cite{Shi2025MemoryVLAPM} maintains a perceptual-cognitive memory. However, these memory-augmented VLAs suffer from a critical limitation: they operate exclusively at the low control frequency ($\sim$1-10Hz) of the visual planner. Because their memory banks rely entirely on low-frequency visual or proprioceptive states, they inherently lack the high-bandwidth physical feedback required for real-time reflexive adjustments, remaining firmly constrained within the ``Slow'' System 2 paradigm.
\subsection{Tactile-Augmented Robotic Manipulation}
Recognizing the limitations of purely visual models in contact-rich tasks, recent works have attempted to integrate tactile and force modalities into VLA architectures. ForceVLA~\cite{yu2025forcevla} introduces a force-aware Mixture-of-Experts module to integrate 6-DoF force feedback during action decoding, while OmniVTLA~\cite{cheng2025omnivtla} employs a semantic-aligned tactile encoder to process multi-modal sensory inputs. Despite these advancements, existing tactile VLAs are fundamentally bottlenecked by their synchronous architectures. Because they feed tactile or force tokens directly into the heavy VLA backbone, the sensory processing is strictly constrained by the VLA's low inference speed. Consequently, these models can only utilize instantaneous tactile snapshots at a drastically reduced frequency, failing to maintain a continuous, high-frequency historical memory. This lossy integration discards critical high-frequency transients (e.g., sudden micro-slips or the snap-through vibrations of a button). In contrast, TacMamba completely decouples the sensory reflex from visual planning, enabling continuous, lossless encoding of 100Hz tactile history without being bottlenecked by the low-frequency VLA.
\subsection{Efficient Long-Sequence Modeling}
Modeling extensive historical sequences is a fundamental challenge in deep learning. While Transformers have been adapted for long-term time-series forecasting~\cite{nie2022time,kim2021reversible}, their $\mathcal{O}(N^2)$ latency renders them unviable for high-frequency embedded control loops. Conversely, traditional Recurrent Neural Networks (RNNs) offer fast inference but are notoriously plagued by catastrophic forgetting over thousands of time steps~\cite{pascanu2013difficulty,bai2018empirical}. Recently, Structured State Space Models (SSMs)~\cite{gu2022efficiently} and their selective variant, Mamba~\cite{gu2024mamba}, have emerged as a paradigm shift. By maintaining a compressed hidden state, Mamba achieves linear-time sequence modeling and strict $\mathcal{O}(1)$ inference latency~\cite{ahamed2024timemachine}. Inspired by this, TacMamba pioneers the application of Mamba to high-frequency tactile streams, effectively resolving the memory-latency trade-off and enabling lossless frequency alignment between System 1 and System 2.

\section{HARDWARE SYSTEM}
\begin{figure}[t] 
    \centering
    \includegraphics[width=1.0\linewidth]{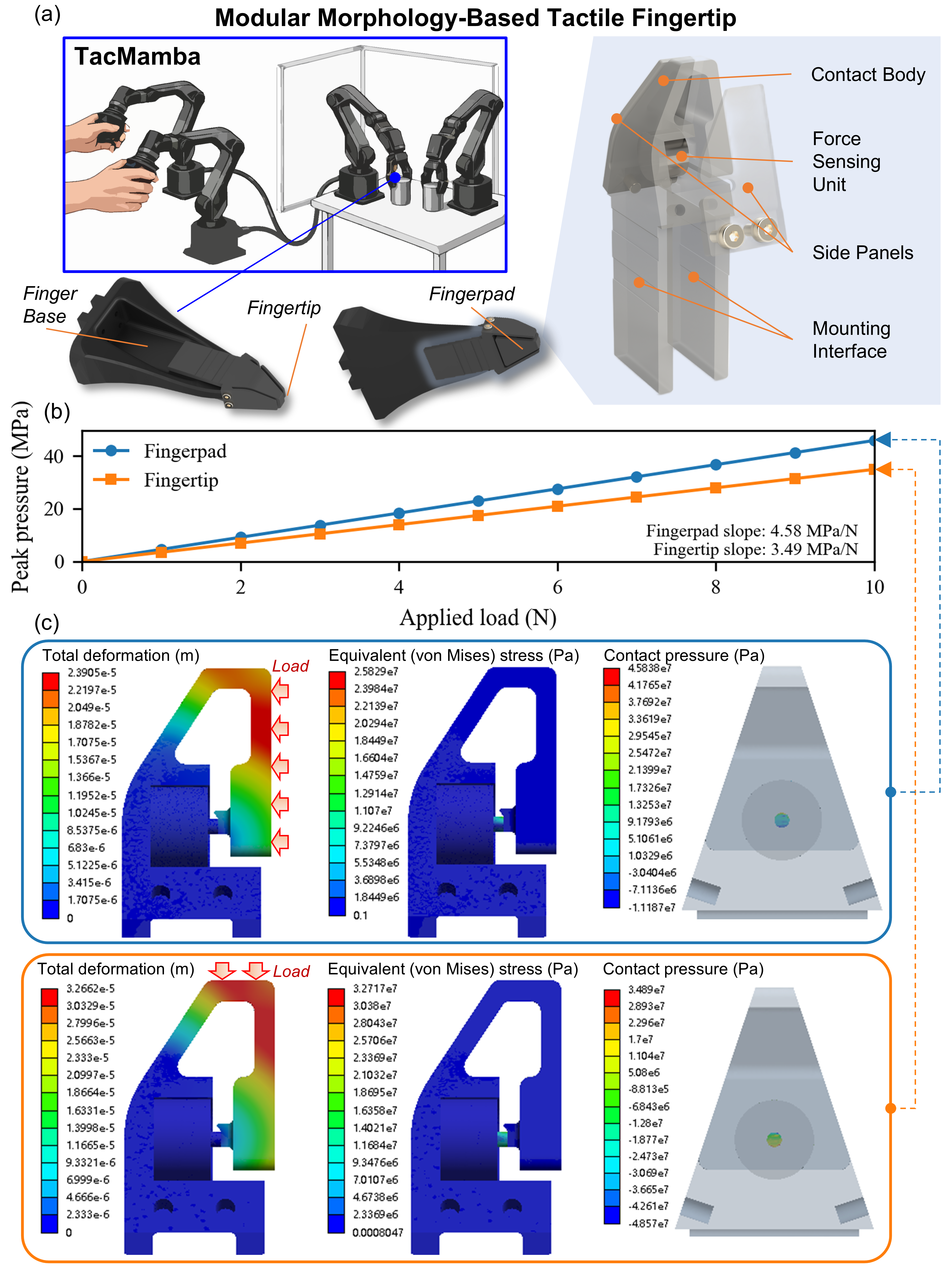}
\caption{\textbf{System Overview.}
(a) Modular morphology-based tactile fingertip design, where an integrated compliant contact body supports both fingertip and fingerpad interactions and mechanically projects distributed contacts onto a single-axis force sensor; lateral side panels protect the internal structure, and a reconfigurable dual-clamp interface enables attachment to generic parallel grippers.
(b) FEM-based characterization of force projection, showing peak contact pressure versus applied load for fingertip and fingerpad contacts, together with (c) representative visualizations of total deformation, equivalent (von Mises) stress, and contact pressure.
}
    \label{fig:system_overview}
\end{figure}

TacMamba is a unified hardware--software system that enables high-frequency tactile reflexes to inform low-frequency VLA reasoning without downsampling, Hardware system overview is shown in (Fig.~\ref{fig:system_overview}). We develop a modular morphology-based tactile fingertip that mechanically projects distributed surface contacts onto a single sensing axis via an integrated compliant contact body,leveraging bio-inspired soft transmission principles~\cite{wettels2008biomimetic}. The module consists of the compliant contact body, a single-axis force sensing unit, lateral side panels, and a reconfigurable dual-clamp mounting interface that attaches to generic parallel grippers and can be adapted to different jaw geometries (Fig.~\ref{fig:system_overview}a). Unlike designs that separate distal and palmar sensing, the fingertip and fingerpad regions are realized as a single continuous compliant structure, allowing both fingertip interactions (e.g., button pressing) and fingerpad interactions (e.g., surface contact and enclosure) to propagate through a continuous force pathway and be mechanically funneled to the force sensor as a unified 1D tactile stream. The side panels shield the internal transmission structure, provide mechanical protection, and constrain undesired lateral deformation during contact~\cite{odhner2014compliant}.

To quantify how spatially distinct contacts are mapped to the sensor interface, we perform finite-element analysis under applied loads from 0 to 10\,N and report total deformation, equivalent (von Mises) stress, and contact pressure on the sensor face (Fig.~\ref{fig:system_overview}c). The resulting pressure--load curves exhibit an approximately linear mapping for both fingertip and fingerpad contacts, with the fingerpad configuration showing a higher pressure gain (Fig.~\ref{fig:system_overview}b). These results confirm that contacts applied at different surface regions are consistently projected onto a single sensing axis through the compliant structure, yielding a predictable and monotonic force transmission pathway.

We emphasize that this design does not aim to recover explicit contact location from tactile sensing alone. Instead, we intentionally adopt a single-axis force sensor as a system-level trade-off. In our framework, coarse contact location is already partially available from visual observations provided by the VLA backbone, whereas critical interaction cues such as click-through, snap-through, and release primarily manifest as high-frequency temporal force variations that are poorly captured by vision. Therefore, a 1D force signal is sufficient to complement visual perception by supplying precise magnitude and dynamic feedback, consistent with findings that aggregated force history contains rich information for complex manipulation~\cite{fazeli2019see}. This minimalist tactile interface aligns with TacMamba’s objective of injecting fast physical reflexes into slow visual reasoning.

\section{METHODOLOGY}
\label{sec:methodology}

TacMamba bridges the frequency gap between high-speed tactile reflexes (System 1) and low-frequency visual planning (System 2). The proposed framework comprises two core components: (A) a hardware-aware Mamba encoder designed for continuous-time tactile modeling, and (B) a dual-stage training strategy that enforces temporal causality through adaptive sampling.

\subsection{Tactile Encoder via Mamba Architecture}
\label{subsec:network_arch}

We adopt the \textbf{Mamba} architecture as our backbone, leveraging its core Selective State Space Model (SSM) mechanism. This allows us to model the continuous physics of contact while handling abrupt dynamic discontinuities (e.g., impacts) through input-dependent selectivity.

\subsubsection{Continuous-Discrete Modeling}
We treat the 1D tactile signal $x(t) \in \mathbb{R}$ as a continuous-time forcing function driving a latent physical system. This system maps the input to a latent state $h(t) \in \mathbb{R}^N$ via a linear time-invariant (LTI) ODE:
\begin{equation}
    \dot{h}(t) = \mathbf{A}h(t) + \mathbf{B}x(t), \quad y(t) = \mathbf{C}h(t)
    \label{eq:continuous_ode}
\end{equation}
To align with the discrete control cycles (time step $\Delta t$) of robotic hardware, we discretize Eq.~\ref{eq:continuous_ode} using the Zero-Order Hold (ZOH) principle, preserving physical consistency:
\begin{equation}
    h_t = \bar{\mathbf{A}} h_{t-1} + \bar{\mathbf{B}} x_t
    \label{eq:recurrence}
\end{equation}
where $\bar{\mathbf{A}} = \exp(\Delta t \mathbf{A})$ and $\bar{\mathbf{B}}$ are the discretized system parameters.

\subsubsection{Capturing Hybrid Dynamics via Selectivity}
Standard LTI systems rely on fixed parameters, which are insufficient for manipulation tasks characterized by hybrid dynamics. Mamba addresses this through its core selective mechanism, which parameterizes the system dynamics as a function of the input: specifically, we compute input-dependent parameters $(\Delta_t, \mathbf{B}_t, \mathbf{C}_t) = \text{Linear}(x_t)$ before discretization. This allows the model to function as an adaptive physical filter, capable of suppressing noise during static grasping while aggressively updating its internal state upon detecting transient contact events like a ``click''.

\subsubsection{Efficient Hierarchical Architecture with $\mathcal{O}(1)$ Inference}
\begin{figure}[t] 
    \centering
    \includegraphics[width=1.0\linewidth]{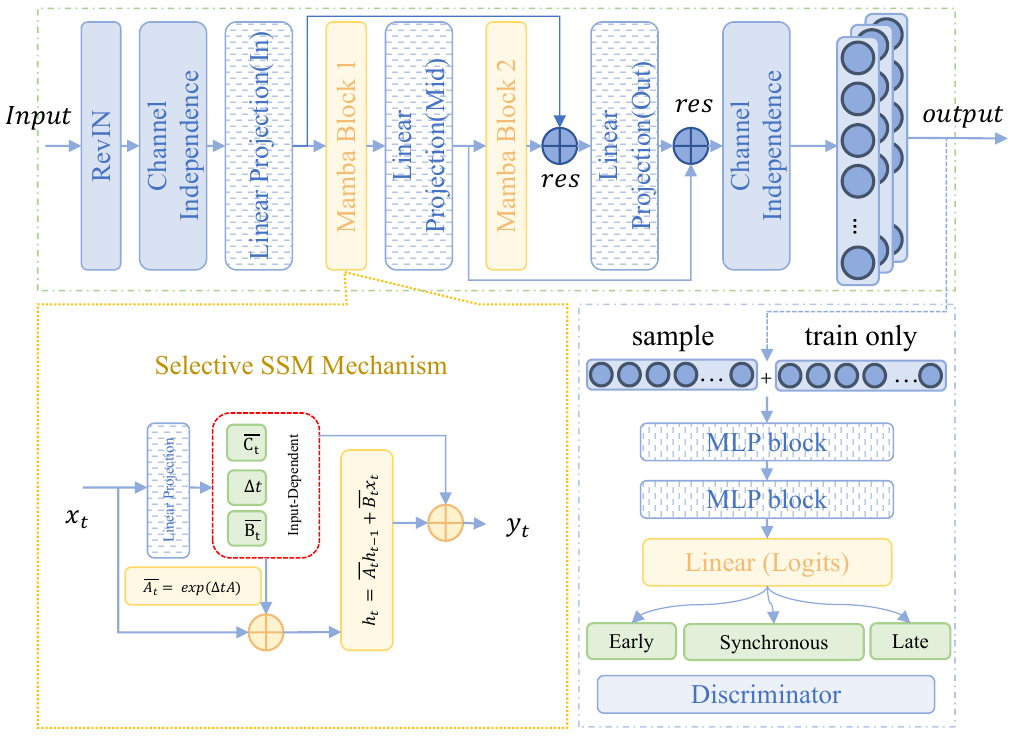}
    \caption{The TacMamba Network Architecture.
    Top: The core TacMamba backbone processes continuous tactile streams via a hierarchical Selective SSM, utilizing RevIN and Channel Independence for feature extraction. 
    Bottom-Left: An expanded view of the Mamba block mechanism, illustrating how input-dependent parameters ($\Delta t, B_t, C_t$) model hybrid dynamics. 
    Bottom-Right: The auxiliary discriminator network, employed \textit{exclusively during the training phase} to facilitate robust feature learning and enforce temporal causality.}
    \label{fig:architecture}
\end{figure}

As illustrated in Fig.~\ref{fig:architecture} (Top), our network processes multivariate non-stationary force signals $\mathbf{X}$. To handle sensor heterogeneity, we briefly apply \texttt{RevIN} and \texttt{Channel Independence} to extract universal physical features. The core backbone consists of stacked Mamba blocks (detailed in Fig.~\ref{fig:architecture}, Bottom-Left) that hierarchically capture features ranging from high-frequency impulses to long-term stability trends. 

The most critical advantage of this architecture is its Constant-Time Inference capability. Exploiting the recurrent formulation in Eq.~\ref{eq:recurrence}, we maintain the compressed hidden state $h_{t-1}$ in memory, making the update for a new force reading $x_t$ strictly $\mathcal{O}(1)$. This stands in sharp contrast to Transformers, ensuring that our ``Fast'' tactile system (100Hz) remains completely decoupled from the latency of the ``Slow'' visual policy.

\subsection{Training Strategy}
\label{subsec:training}

To mitigate the scarcity of labeled tactile data and the inherent temporal sparsity of critical contact events, we propose a robust two-stage protocol. 

\subsubsection{Stage 1: Self-Supervised Pre-training via Temporal Discrimination}
To enable the Mamba encoder to comprehend contact dynamics without manual labels, we design a Ternary Temporal Discrimination task. 

Automatic Segmentation: First, we treat the tactile stream as a sequence of physical sub-tasks. We automatically segment long trajectories into discrete \textit{sub-task phases} based on the tactile gradient magnitude $\|\nabla x_t\|$, specifically detecting transitions from non-contact to contact. 

Objective: Based on these phases, we generate state pairs $(h_i, h_j)$ and train a discriminator $D$ to classify their temporal relationship into three categories: $y_{ij} \in \{0: t_i < t_j, 1: t_i \approx t_j, 2: t_i > t_j\}$. 
Crucially, we employ a Relation-Balanced Sampling strategy to rigorously enforce a uniform distribution across these three classes (Early, Simultaneous, Late). This prevents the model from collapsing into trivial guessing strategies (e.g., always predicting "future") and forces the encoder to embed the causal structure of physical interactions into the hidden state space.

We train the discriminator by minimizing the standard Cross-Entropy loss:
\begin{equation}
    \mathcal{L}_{pre} = -\mathbb{E}_{(h_i, h_j) \sim \mathcal{D}_{bal}} \sum_{c=0}^{2} \mathbb{I}(y_{ij}=c) \log D(h_i, h_j)_c
\end{equation}

\subsubsection{Stage 2: Tactile-Guided VLA Tuning with Phase Sampling}
In this stage, we freeze the pre-trained Mamba encoder and project its compressed state $h_t$ as a ``Soft Prompt'' $z_{tac}$ into the VLA. 

Phase-Uniform Sampling: A critical challenge in robotic manipulation is that key interaction frames (e.g., contact transition) are extremely sparse compared to long-duration idle modes. As shown in the sparse force peaks of Fig.~\ref{fig:system_arch}, critical events occupy a tiny fraction of the timeline.

To address this, we utilize the automatic segmentation from Stage 1 to decompose each trajectory into a sequence of $K$ discrete sub-task phases $\mathcal{P} = \{P_1, P_2, \dots, P_K\}$.
Standard uniform sampling ($p(t) \propto 1/T$) biases the model towards long-duration phases (e.g., static holding). In contrast, we employ a Phase-Uniform Sampling strategy that enforces a uniform distribution across phases:
\begin{equation}
    p(t \in P_k) \propto \frac{1}{K \cdot |P_k|}
\end{equation}
This ensures that each sub-task mode contributes equally to the training batch regardless of its temporal duration. Consequently, transient but critical phases (like the instant of snap-through) are weighted equally with stable modes, ensuring the VLA learns the complete manipulation logic without ignoring short-duration tactile cues.

\section{EXPERIMENTS}
\label{sec:experiments}
\subsection{Experimental Setup}

Hardware Platform: We utilize an AgileX PiPER lightweight 6-DoF manipulator equipped with a parallel gripper. To capture contact dynamics, a custom single-point high-frequency force sensor is integrated into the fingertip, streaming 1D force data at 100Hz.
This section aims to answer two core research questions:\\
\noindent (1) Efficiency \& Scalability: Can TacMamba maintain real-time responsiveness for high-frequency closed-loop control while modeling ultra-long tactile histories?\\
(2) Robustness under Ambiguity: Does the injection of high-frequency tactile history significantly improve manipulation success rates in scenarios with visual occlusion or state ambiguity?\\
Baselines: We compare our method against two categories of baselines:
\begin{enumerate}
    \item \textit{Backbone Baselines:} For the tactile encoder, we compare TacMamba against 1D-CNN (local convolution), LSTM (recurrent), and Transformer (global attention) to evaluate history compression capabilities.
    \item \textit{System Baseline:} For the end-to-end policy, we compare against \textbf{$\pi_{0.5}$}~\cite{black2025pi}, a state-of-the-art generalist Vision-Language-Action (VLA) model, to assess the value of specialized tactile memory.
\end{enumerate}
\subsection{Performance Evaluation: Accuracy \& Efficiency}
\label{subsec:eval_main}

\begin{figure}[t]
    \centering
    \includegraphics[width=\linewidth]{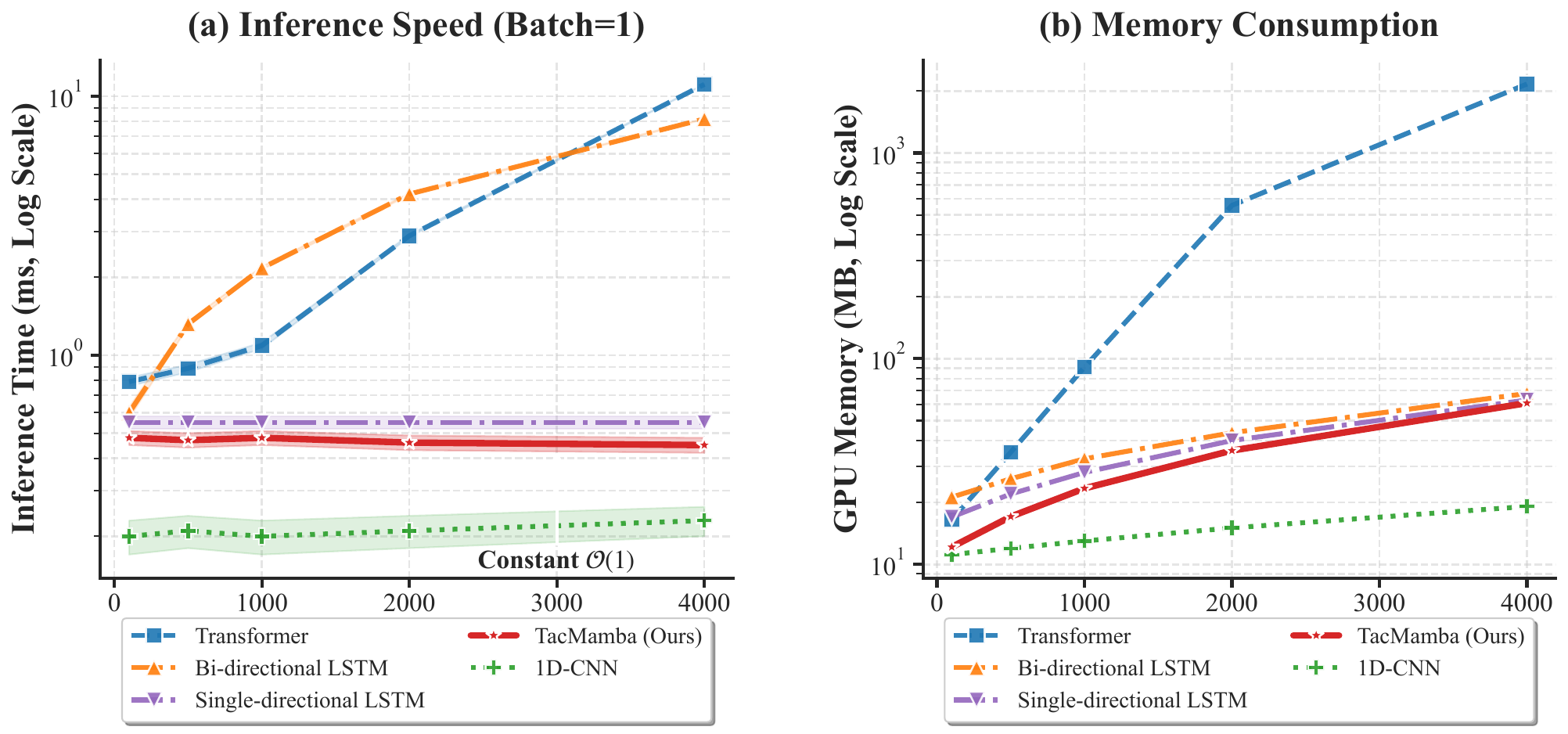}
    \caption{\textbf{Efficiency Analysis.} Inference latency and memory growth.}
    \label{fig:efficiency}
    
    \vspace{2pt} 
    \makeatletter\def\@captype{table}\makeatother 
    \caption{Comparison on the Augmented Real-World Dataset}
    \label{tab:accuracy_results}
    \centering
    \resizebox{0.8\linewidth}{!}{ 
        \setlength{\tabcolsep}{1.5pt}
        \begin{tabular}{lcccc}
            \toprule
            \textbf{Model} & \textbf{Acc.}  & \textbf{Lat.} (ms)  & \textbf{Mem.} (MB)  \\
            \midrule
            1D-CNN & 56.25\% & \textbf{$0.23\pm0.02$} & 19.14 & \\
            Transformer & 81.25\% & $11.37\pm0.17$ & 2164.71 &  \\
            Bi-LSTM & 82.50\% & $9.20\pm0.13$ & 67.79 &  \\
            Single-LSTM & 65.33\% & $0.54\pm0.01$ & 63.70 &  \\
            \midrule
            \textbf{TacMamba} & \textbf{88.89\%} & $0.45\pm0.02$ & 60.85 &  \\
            \bottomrule
        \end{tabular}
    }
    \vspace{-15pt}
\end{figure}

We benchmark TacMamba against standard architectures on the exact same dataset. The quantitative results are summarized in Fig.~\ref{fig:efficiency} and Table~\ref{tab:accuracy_results}.

\subsubsection{Comprehensive Analysis: Accuracy-Efficiency Trade-off}
We present a holistic evaluation of the trade-off between perception fidelity and real-time feasibility on the augmented real-world dataset. The comparison reveals distinct characteristics for each architecture:

\begin{figure*}[t]
    \centering
    \begin{minipage}{0.48\linewidth}
        \centering
        \includegraphics[width=\linewidth]{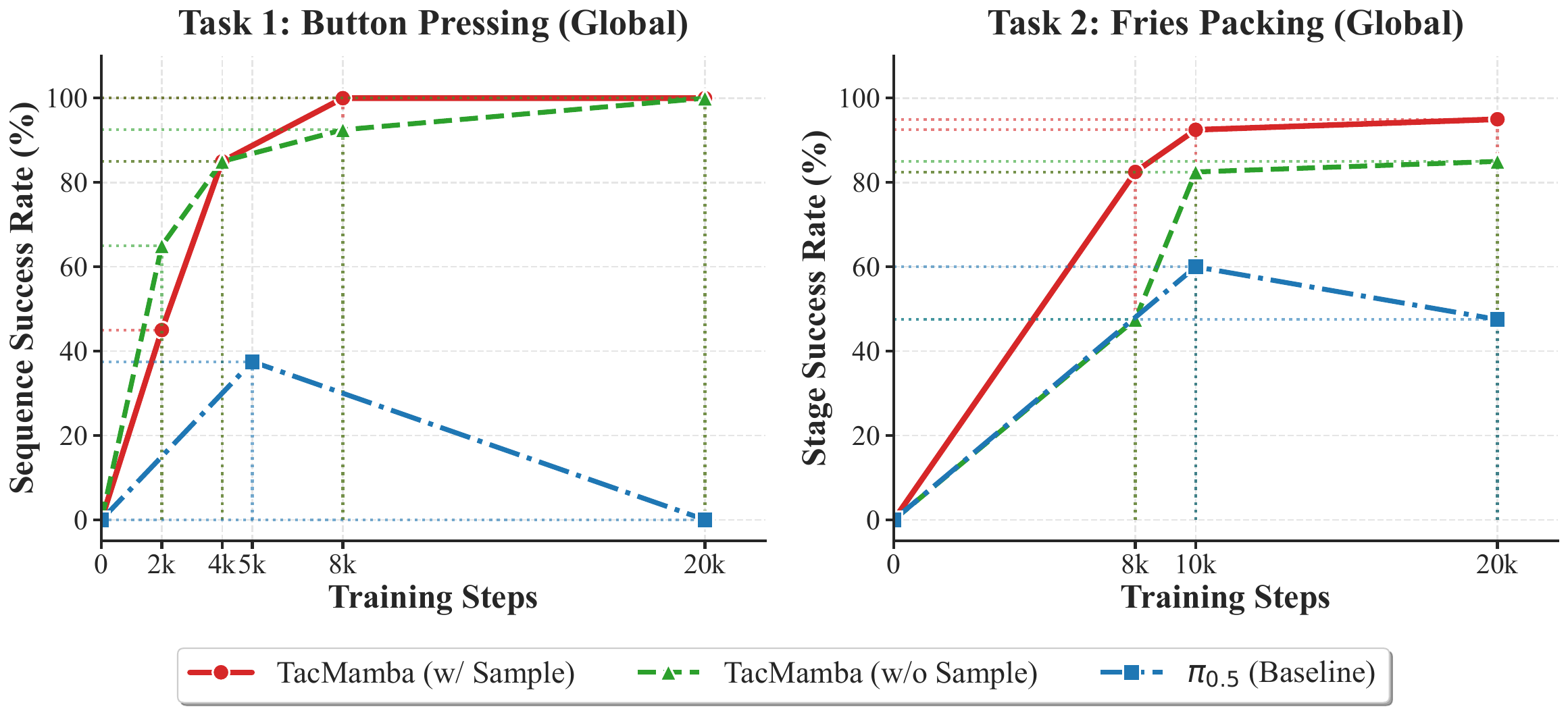}
        \caption{\textbf{Global Task Success Rate.} Comparison of full-task completion rates over training steps. TacMamba (Red) achieves rapid convergence and high robustness, while $\pi_{0.5}$ (Blue) suffers from catastrophic failure in the button task due to static frame overfitting.}
        \label{fig:global_success}
    \end{minipage}
    \hfill
    \begin{minipage}{0.48\linewidth}
        \centering
        \includegraphics[width=\linewidth]{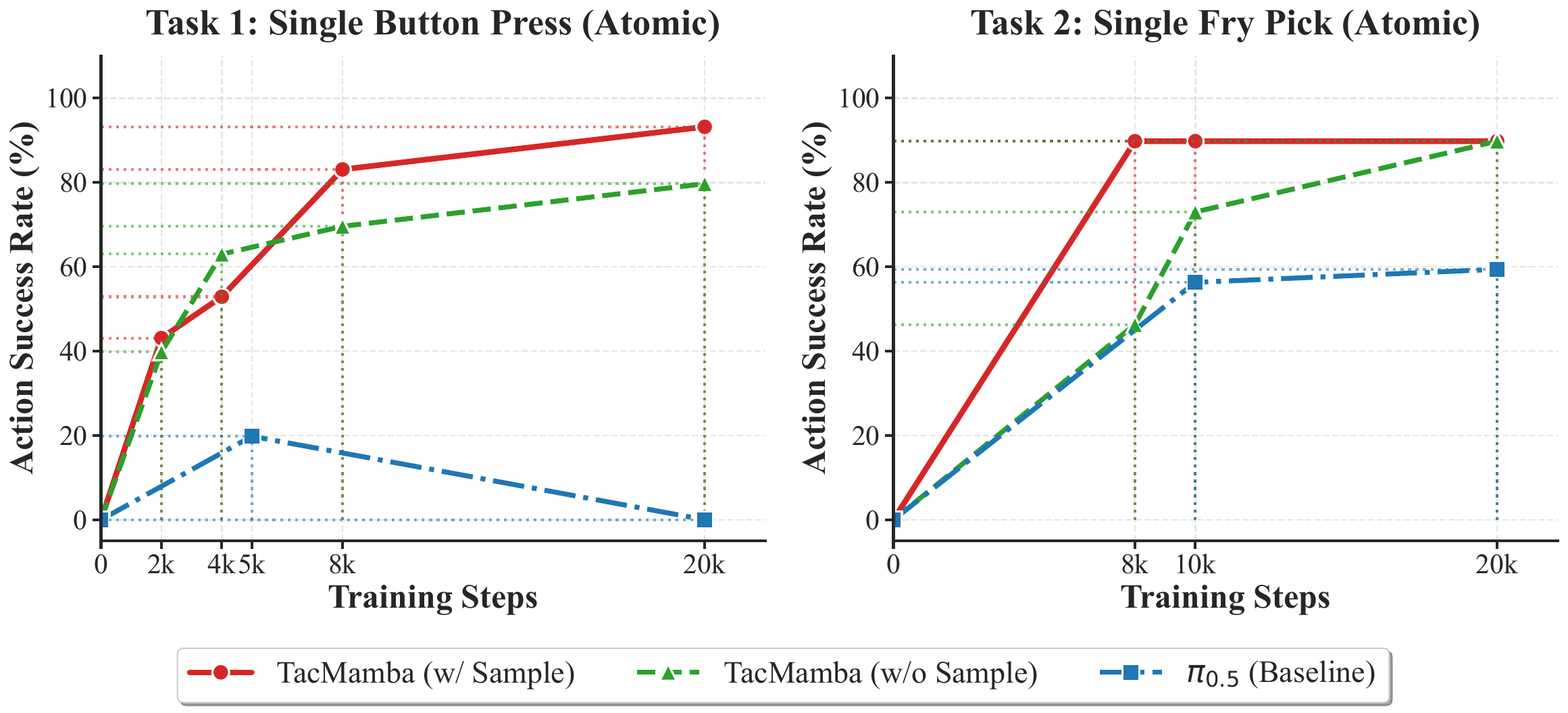}
        \caption{\textbf{Atomic Action Success Rate.} Evaluation of fine-grained interaction quality (e.g., successful single button press without overshoot). TacMamba maintains high physical precision, whereas the baseline struggles with hallucinations and execution errors.}
        \label{fig:atomic_success}
    \end{minipage}
    \vspace{-10pt}
\end{figure*}

TacMamba (Ours): TacMamba achieves \textbf{SOTA accuracy (88.89\%)} with an ultra-low inference latency of 0.45 ms. Its $\mathcal{O}(1)$ complexity offers a 21$\times$ speedup over the LSTM, uniquely satisfying high-frequency control requirements ($>100$Hz) without sacrificing precision.

RNNs (LSTM): The Bi-LSTM achieves 82.50\% accuracy but requires re-evaluating the entire history at each step, resulting in a 9.20 ms latency that breaches strict streaming constraints. The Single-LSTM is fast (0.54 ms) but suffers from catastrophic forgetting over high-frequency sequences, degrading accuracy to 65.33\%.

Transformer: The Transformer achieves only 81.25\% accuracy due to data inefficiency in extracting features from noisy tactile streams. Furthermore, its $\mathcal{O}(L^2)$ complexity spikes latency to 11.37 ms with excessive memory usage, rendering it infeasible for edge computing.

1D-CNN: While computationally efficient (0.23 ms), the 1D-CNN exhibits the lowest accuracy (56.25\%). Its limited receptive field fails to contextualize local vibrations within the broader task history, leading to state confusion.

\subsection{Real-World Validation: Sequential Interaction under Occlusion}
\label{subsec:real_world}

To comprehensively validate performance in physical environments and explicitly evaluate the limitations of traditional visual models in long-horizon scenarios, we designed three distinct manipulation tasks---``Sequential Button Pressing'', ``Blind French Fry Packing'', and ``Deformable Clothes Folding''. These tasks are formulated to rigorously benchmark TacMamba against the visual-only $\pi_{0.5}$ baseline across varying degrees of visual occlusion and state ambiguity.

Baseline Selection Strategy: While we performed extensive comparisons with LSTM and Transformer architectures on the offline dataset (Sec.~\ref{subsec:eval_main}), we exclude them from these real-world closed-loop tasks. A critical constraint for System 1 reflexes is the 100Hz (10 ms) control cycle. However, the inference latency of both Transformer (quadratic complexity) and LSTM (sequential processing) scales with the input history length. As the tactile history grows during long-horizon manipulation, their computation time inevitably breaches this strict real-time limit, introducing dangerous delays. Consequently, TacMamba, with its constant-time $\mathcal{O}(1)$ inference, is the only architecture capable of sustaining the high-frequency control loop, making $\pi_{0.5}$ (visual-only) the only viable system-level baseline.

\subsubsection{Task Definition}

Task 1: Sequential Button Pressing. The robot is required to press three buttons in a specific color sequence (Blue $\rightarrow$ Pink $\rightarrow$ Purple). This task introduces three critical challenges:

\begin{enumerate}
    \item Visual Occlusion: The robot's end-effector completely occludes the target area during operation, rendering the $\pi_{0.5}$ visual encoder blind.
    \item State Ambiguity: The button trigger state (``Click'') provides only subtle vibration feedback, which is visually imperceptible.
    \item Perceptual Aliasing (Memory Gap): Since the buttons appear identical and the visual feed remains static due to occlusion, a model relying solely on current observation cannot distinguish the sequence phase (e.g., the first press vs. the second). Success requires persistent memory of past interactions.
\end{enumerate}

Task 2: Blind French Fry Packing. The robot must sequentially pick three french fries from a plate and place them into an opaque box. The core difficulty lies in Cumulative State Estimation:\\
 Since the box's interior is occluded and external vision cannot penetrate it, the robot is effectively ``blind'' to the task progress. The system must rely entirely on long-horizon tactile history to track counting (i.e., remembering previous successful picks) to determine when to terminate the task.

\begin{figure*}[t] 
    \centering
    \includegraphics[width=1.0\linewidth]{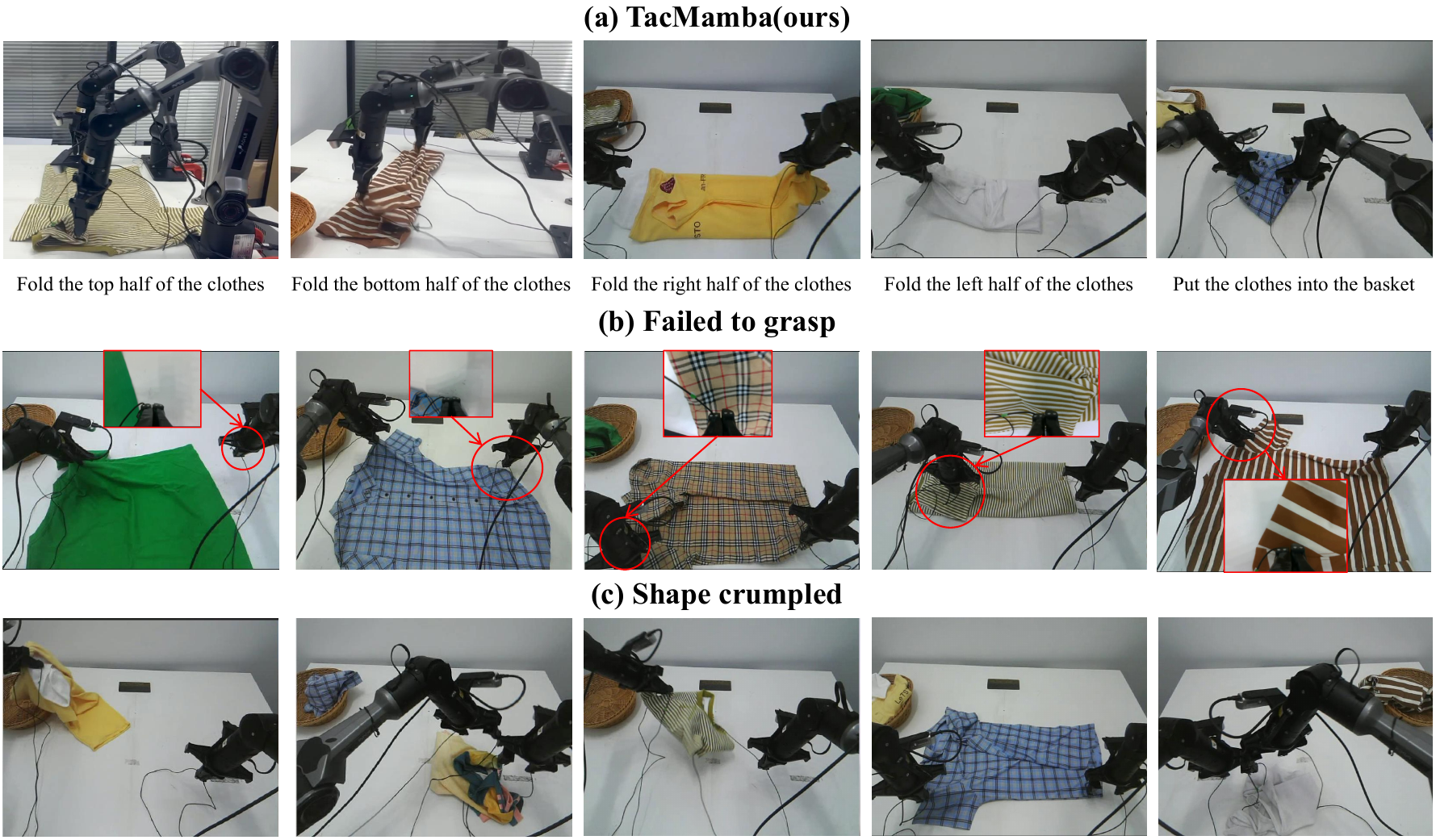} 
    \caption{\textbf{Qualitative folding results comparison.} (a) Successful cloth folding workflow executed by the full TacMamba model. The zoomed-in views highlight precise and stable grasps on the fabric corners enabled by high-frequency tactile feedback. (b) and (c) Typical failure modes observed when the phase-prioritized sampling strategy is disabled (ablation): "Failed to grasp" (missing the corner due to visual ambiguity) and "Shape crumpled" (improper tension control causing deformation).}
    \label{fig:folding_qualitative_comparison}
\end{figure*}

Task 3: Deformable Clothes Folding. As depicted in Fig.~\ref{fig:folding_qualitative_comparison}, the robot executes a five-step workflow: folding the garment across four directions (top, bottom, right, left) and placing it into a basket. This extended deformable manipulation rigorously tests VLA models against three core challenges:

\begin{enumerate}
    \item Precise Control: The fabric's extreme thinness demands fine-grained, continuous physical interaction.
    \item Visual Ambiguity: Dynamic wrinkles easily deceive purely visual models, making tactile feedback indispensable to confirm a secure ``pinch''.
    \item Long-Horizon Overfitting: The prolonged sequence exacerbates state forgetting and policy overfitting to static non-contact frames, degrading execution quality over time.
\end{enumerate}

Evaluation Metrics: To provide a granular analysis of physical interaction quality, we report performance on two levels:

\begin{enumerate}
    \item Global Task Success: The percentage of episodes where the robot completes the entire long-horizon sequence (e.g., all 3 buttons pressed in correct order, or 3 fries packed).
    \item Atomic Action Success: Strict criteria to rule out ``lucky'' sub-task successes. \textit{For Button Pressing:} Must trigger the tactile ``click'' and halt immediately (overshoot or phantom presses are failures). \textit{For Fries Packing:} Must firmly grasp, transport, and release the fry into the box (grasping air, dropping, or edge-snagging are failures).
\end{enumerate}

\subsubsection{Results Comparison and Analysis}

We evaluate TacMamba (with and without our sampling strategy) against $\pi_{0.5}$ across different training stages. 

To ensure statistical significance and eliminate stochasticity, we adopted a rigorous evaluation protocol: for each training checkpoint, we conducted experiments across 10 randomized initial positions, with 5 independent trials per position (totaling 50 evaluation episodes per data point).

\begin{table}[h]
    \centering
    \caption{Performance on Sequential Button Pressing Task}
    \label{tab:button_baselines}
    \resizebox{0.85\linewidth}{!}{
        \begin{tabular}{lc}
            \toprule
            Method & Success Rate \\
            \midrule
            \textit{External Visual Baselines$^*$:} & \\
            OpenVLA~\cite{kim2024openvla} & 6\% \\
            CogACT~\cite{li2024cogact} & 15\% \\
            MemoryVLA~\cite{Shi2025MemoryVLAPM} & 58\% \\
            \midrule
            \textit{Our Implementations:} & \\
            $\pi_{0.5}$~\cite{black2025pi} (Early stage, 5k steps) & 37.5\% \\
            $\pi_{0.5}$~\cite{black2025pi} (Converged, 20k steps) & 0\% \\
            \textbf{TacMamba (Phase-Uniform, Ours, 20k steps)} & \textbf{100\%} \\
            \bottomrule
        \end{tabular}
    }
    \vspace{3pt}
    
    \raggedright \footnotesize 
    $^*$Results for external visual baselines are directly sourced from~\cite{Shi2025MemoryVLAPM}. Their experimental setup is identical to ours, with the sole exception of utilizing a Franka Panda manipulator instead of our AgileX PiPER.
    \vspace{-10pt}
\end{table}

1. Baseline Limitations and Temporal Overfitting to Static Frames: The pure visual baseline ($\pi_{0.5}$) struggles in long-horizon tasks, as tracked in Fig.~\ref{fig:global_success}. In the \textit{Sequential Button Pressing} task, a temporal degradation is observed: the baseline achieves 37.5\% success at 5k steps, but drops to 0\% by 20k steps. This performance regression exposes the vulnerability of vision-only VLA models to temporal overfitting caused by data sampling bias. Because standard random sampling over-represents static, non-contact frames (which dominate the long-horizon sequence), the visual policy overfits to these stationary features over time. Consequently, the converged model tends to hover above the button, losing the ability to execute the dynamic pressing action. 
This degradation is not an isolated issue; as summarized in Table~\ref{tab:button_baselines}, visual memory architectures like MemoryVLA~\cite{Shi2025MemoryVLAPM} plateau at a 58\% success rate on similar tasks. This performance ceiling stems from the low-frequency nature of visual observations: micro-scale physical events often fall into the blind spots between camera frames, causing state confusion. In contrast, TacMamba overcomes this static-frame overfitting, maintaining a 100\% success rate upon convergence by anchoring its policy on high-frequency tactile reflexes. Similarly, in the Fries Packing task, visual occlusion limits $\pi_{0.5}$ to a 47.5\% global success rate.

2. TacMamba's High-Frequency Robustness: TacMamba effectively bypasses these perceptual bottlenecks. As demonstrated in both the global success (Fig.~\ref{fig:global_success}) and atomic action success (Fig.~\ref{fig:atomic_success}) curves, our full model maintains exceptional stability. By processing 1D tactile streams at 100Hz, it reliably captures transient events and maintains strict physical precision. TacMamba achieves 100\% success in the button task by 8k steps and reaches 95\% in the fries task, successfully performing implicit counting via long-horizon tactile history without visual confirmation.

3. Ablation on Phase-Uniform Sampling: Comparing TacMamba trained with our Phase-Uniform Sampling against a Standard Uniform variant highlights the strategy's data efficiency and fine-grained control capabilities. As explicitly tracked by the atomic action success rate in Fig.~\ref{fig:atomic_success}, the phase-uniform approach significantly reduces micro-level execution errors (e.g., overshoot or phantom grasps). Consequently, at 8k steps in the fries task, it yields an 82.5\% global success rate, nearly doubling the uniform baseline (47.5\%). By explicitly prioritizing sparse, high-frequency contact events, our strategy improves overall sample efficiency by approximately 2.5$\times$.

4. Deformable Manipulation \& Hardware Constraints: Due to the fabric's extreme thinness, unconverged policies often apply excessive downward force, risking physical damage to the custom tactile sensor against the rigid tabletop. Therefore, we evaluated the \textit{Clothes Folding} task exclusively at full convergence (20k steps). While all models achieved 100\% macro-folding success, fine-grained quality varied. The visual baseline and the uniform-sampling ablation suffered from micro-failures (e.g., ``Failed to grasp'' or ``Shape crumpled'') due to ambiguous visual wrinkles (Fig.~\ref{fig:folding_qualitative_comparison}). The full TacMamba model maintained robust reflexes, reliably pinching exactly when the fabric edge was felt.

\section{CONCLUSIONS}

In this work, we presented TacMamba, a hardware-software architecture that successfully bridges the critical spatiotemporal gap between high-frequency tactile reflexes and low-frequency visual reasoning. By formulating continuous 1D tactile streams through a Mamba-based architecture, our approach achieves SOTA accuracy (88.89\%) with strict $\mathcal{O}(1)$ inference latency (0.45 ms). This outperforms Transformer baselines by a 25$\times$ speedup and completely bypasses their latency bottlenecks, representing the first architecture capable of processing ultra-long tactile histories ($L=5000$) within the hard real-time constraints of embedded hardware. Furthermore, it overcomes the catastrophic forgetting inherent in standard LSTMs, enabling lossless, real-time processing of micro-scale contact dynamics.

Additionally, our phase-uniform sampling strategy effectively mitigates the data imbalance of sparse contact events, drastically improving sample efficiency during VLA fine-tuning. Extensive physical evaluations confirm that TacMamba entirely overcomes the temporal overfitting to static frames that causes purely visual baselines (e.g., $\pi_{0.5}$) to fail. By strictly satisfying 100Hz control loop constraints and maintaining fine-grained physical precision without overshoot, our system achieves robust task execution across visually occluded and long-horizon scenarios, reliably solving complex challenges like sequential button pressing, implicit state counting, and deformable clothes folding.

Despite these advances, our system has certain limitations. Our current setup utilizes a 1D force sensor rather than high-resolution vision-based tactile sensors (e.g., GelSight). While sufficient for isolating high-frequency temporal dynamics and avoiding the computational overhead of image processing~\cite{Visual-tactile}, this prevents the extraction of fine-grained spatial topologies. Additionally, rather than exhaustive large-scale benchmarking, our physical evaluation deliberately targets specific edge cases (e.g., severe occlusion), comparing primarily against the state-of-the-art $\pi_{0.5}$~\cite{black2025pi}. Nevertheless, TacMamba's decoupled nature naturally supports plug-and-play integration with broader open-source VLA frameworks.

Future work will address these constraints by integrating high-dimensional GelSight image sequences directly into the Mamba network for spatially-rich contact tasks, and investigate its potential to facilitate zero-shot sim-to-real transfer for multi-fingered dexterous hands.




\balance
\bibliographystyle{IEEEtran}
\bibliography{references}
\end{document}